\documentclass[10pt,twocolumn,letterpaper]{article}

\usepackage{xspace}
\usepackage{times}
\usepackage{epsfig}
\usepackage{graphicx}
\usepackage{amsmath}
\usepackage{amssymb}
\usepackage{booktabs}
\usepackage[numbers]{natbib}
\usepackage{xcolor}
\usepackage{balance}

\newif\ifarxiv
\newif\ifnotarxiv

\arxivtrue
\notarxivfalse

\usepackage[pagebackref=true,breaklinks=true,letterpaper=true,colorlinks,bookmarks=false]{hyperref}

\date{}
\makeatletter
\DeclareRobustCommand\onedot{\futurelet\@let@token\@onedot}
\def\@onedot{\ifx\@let@token.\else.\null\fi\xspace}

\def\eg{\emph{e.g}\onedot} 
\def\ie{\emph{i.e}\onedot}

\def\wrt{w.r.t\onedot} 
\def\etal{\emph{et al}\onedot}
\makeatother

\def \pzo {\phantom{0}} 
\def \dzo {\phantom{00}}
\def \tzo {\phantom{000}}
\def \qzo {\phantom{0000}}

\newcommand{\stdminus}[1]{\scalebox{0.65}{$\pm #1$}}
\newcommand{\rv}[1]{{\color{red}#1}}
\newcommand{\hugo}[1]{{\color{blue!20!red}[\textbf{Hugo}:#1]}}
\newcommand{\matthijs}[1]{{\color{blue}[\textbf{Matthijs}:#1]}}
\newcommand{\ben}[1]{{\color{orange}[\textbf{Benjamin}:#1]}}
\newcommand{\alaa}[1]{{\color{green!40!red}[\textbf{Alaa}:#1]}}
\newcommand{\pierre}[1]{{\color{green!20!red}[\textbf{Pierre}:#1]}}
\newcommand{\aj}[1]{{\color{green!20!red}[\textbf{Armand}:#1]}}

\makeatletter
\renewcommand{\paragraph}{%
  \@startsection{paragraph}{4}%
  {\z@}{1.0ex \@plus 1ex \@minus .2ex}{-1em}%
  {\normalfont\normalsize\bfseries}%
}
\makeatother

\usepackage{enumitem}
\setlist[itemize]{%
topsep=5pt,
labelsep=5pt,%
labelindent=0.4\parindent,%
itemindent=0pt,%
leftmargin=*,%
itemsep=-1pt 
}

\begin{document}

\title{LeViT: a Vision Transformer in ConvNet's Clothing \\ for Faster Inference}

\author{
Benjamin Graham \and
Alaaeldin El-Nouby \and
Hugo Touvron \and
Pierre Stock  \and 
Armand Joulin \and
Herv\'e J\'egou \and
Matthijs Douze
}

\maketitle
\begin{abstract}
We design a family of image classification architectures that optimize the trade-off between accuracy and efficiency in a high-speed regime. 
Our work exploits recent findings in attention-based architectures, which are competitive on highly parallel processing hardware. We revisit principles from the extensive literature on convolutional neural networks to apply them to transformers, in particular activation maps with decreasing resolutions. 
We also introduce the attention bias, a new way to integrate positional information in vision transformers.

As a result, we propose LeVIT: a hybrid neural network for fast inference image classification. 
We consider different measures of efficiency on different hardware platforms, so as to best reflect a wide range of application scenarios. 
Our extensive experiments empirically validate our technical choices and show they are suitable to most architectures. Overall, LeViT significantly outperforms existing convnets and vision transformers with respect to the speed/accuracy tradeoff.
For example, at 80\% ImageNet top-1 accuracy, LeViT is 5 times faster than EfficientNet on CPU. We release the code at \url{https://github.com/facebookresearch/LeViT}.
\end{abstract}

\vspace{-10pt}
\section{Introduction}
\label{sec:introduction}

Transformer neural networks were initially introduced for Natural Language Processing applications~\cite{Vaswani2017AttentionIA}. They now dominate in most applications of this field. 
They manipulate variable-size sequences of token embeddings that are fed to a residual architecture. 
The model comprises two sorts for residual blocks: Multi-Layer Perceptron (MLP) and an original type of layer: the self-attention, which allows all pairs of tokens in the input to be combined via a bilinear function. 
This is in contrast to 1D convolutional approaches that are limited to a fixed-size neighborhood.

\begin{figure}
\hspace*{-7mm}\includegraphics[width=1.2\linewidth]{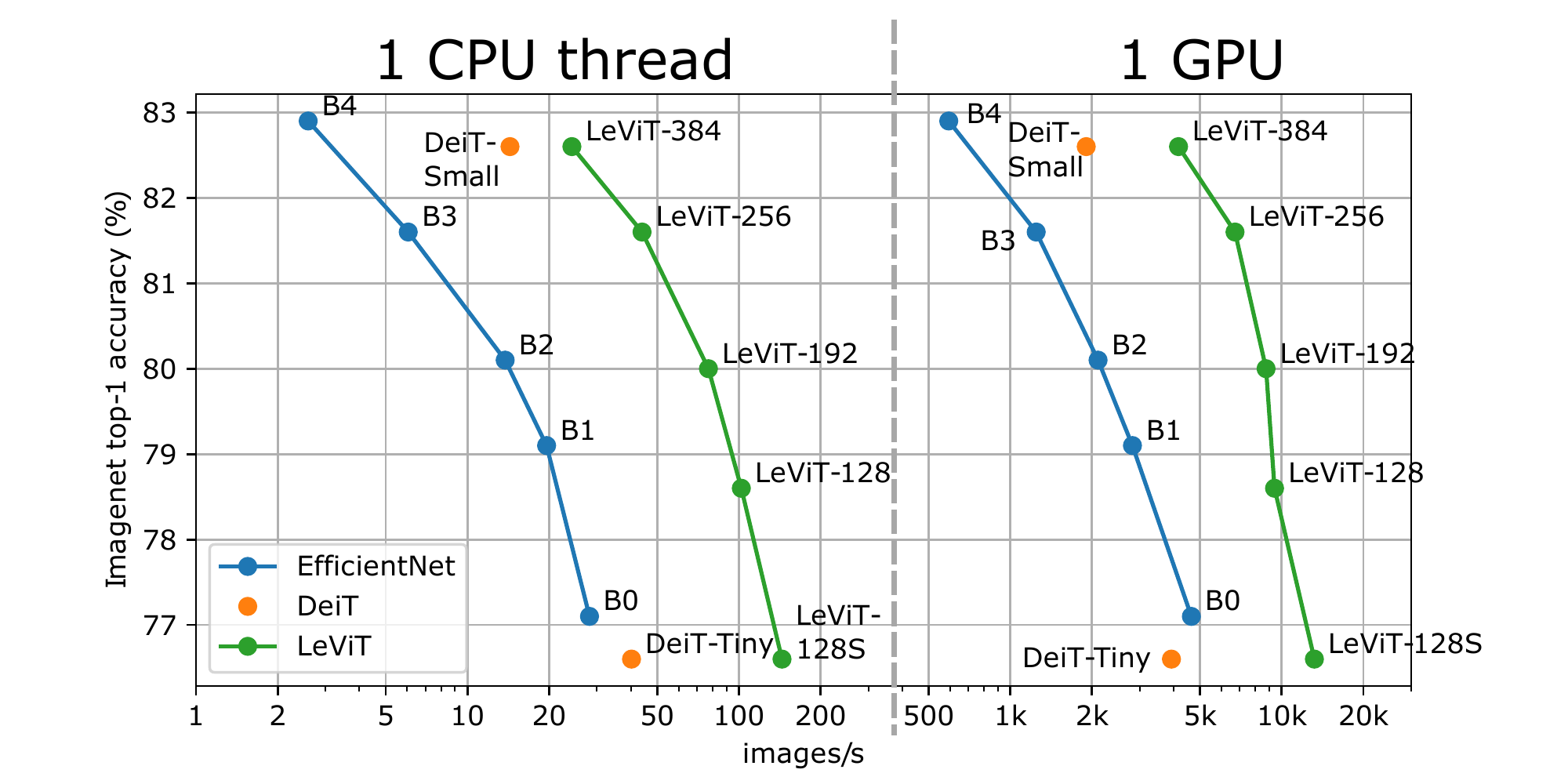}
\caption{\label{fig:mainspeedaccuracyplot}
	Speed-accuracy operating points for convolutional and visual transformers. 
	\emph{Left} plots: on 1 CPU core, \emph{Right:} on 1 GPU.
	 LeViT is a stack of transformer blocks, with pooling steps to reduce the resolution of the activation maps as in classical convolutional architectures. 
}
\end{figure}

Recently, the vision transformer (ViT) architecture~\cite{dosovitskiy2020image} obtained state-of-the-art results for image classification in the speed-accuracy tradeoff with pre-training on large scale dataset. The Data-efficient Image Transformer \cite{Touvron2021DeiT} obtains competitive performance when training the ViT models only on ImageNet \cite{deng2009imagenet}. It also introduces smaller models adapted for high-throughput  inference.

In this paper, we explore the design space to offer even better trade-offs than ViT/DeiT models in the regime of small and medium-sized architectures. We are especially interested in optimizing the performance--accuracy trade-off, such as the throughput (images/second) performance depicted in Figure~\ref{fig:mainspeedaccuracyplot} for Imagenet-1k-val~\cite{Russakovsky2015ImageNet12}. 

While many works \cite{Han2015DeepCC,courbariaux2016binaryconnect,zhou2018dorefanet,wang2019haq,stock2020bit} aim at reducing the memory footprint of classifiers and feature extractors, inference speed is equally important, with high throughput corresponding to better energy efficiency.
In this work, our goal is to develop a Vision Transformer-based family of models with better inference speed on both highly-parallel architectures like GPU, regular Intel CPUs, and ARM hardware commonly found in mobile devices. 
Our solution re-introduces convolutional components in place of transformer components that learn convolutional-like features. 
In particular, we replace the uniform structure of a Transformer by a pyramid with pooling, similar to the LeNet~\cite{lecun1989backpropagation} architecture. Hence we call it LeViT. 

There are compelling reasons why transformers are faster than convolutional architectures for a given computational complexity. 
Most hardware accelerators (GPUs, TPUs) are optimized to perform large matrix multiplications.
In transformers, attention and MLP blocks rely mainly on these operations. 
Convolutions, in contrast, require complex data access patterns, so their operation is often IO-bound.
These considerations are important for our exploration of the speed/accuracy tradeoff.  

\paragraph{The contributions}
of this paper are techniques that allow ViT models to be shrunk down, both in terms of the width and spatial resolution:

\begin{itemize}
\item
	A multi-stage transformer architecture using attention as a downsampling mechanism;
\item
    A computationally efficient patch descriptor that shrinks the number of features in the first layers;
\item
    A learnt, per-head translation-invariant attention bias that replaces ViT's positional embeddding;
\item
    A redesigned Attention-MLP block that improves the network capacity for a given compute time.
\end{itemize}

\section{Related work}
\label{sec:related}

The convolutional networks descended from LeNet~\cite{lecun1989backpropagation} have evolved substantially over time~\cite{Krizhevsky2012AlexNet,Simonyan2015VGG,He2016ResNet,Szegedy2016InceptionResNetV2,Xie2017AggregatedRT,tan2019efficientnet}.
The most recent families of architectures focus on finding a good trade-off between efficiency and performance~\cite{Radosavovic2020RegNet,tan2019efficientnet,Howard2019SearchingFM}.
For instance, the EfficientNet~\cite{tan2019efficientnet} family was discovered by carefully designing individual components followed by hyper-parameters search under a FLOPs constraint.

\paragraph{Transformers.} 
The transformer architecture was first introduced by Vaswani \etal~\cite{Vaswani2017AttentionIA} for machine translation. Transformer encoders primarily rely on 
the self-attention operation in conjunction with feed-forward layers, providing a strong and explicit method for learning long range dependencies. Transformers have been subsequently adopted for NLP tasks providing state-of-the-art performance on various benchmarks \cite{devlin2018bert, Radford2018improving}. 
There have been many attempts at adapting the transformer architecture to images~\cite{parmar2018image,child2019generating}, first by applying them on pixels.
Due to the quadratic computational complexity and number of parameters involved by attention mechanisms, most authors~\cite{child2019generating,Cordonnier2020OnTR} initially considered images of small sizes like in CIFAR or Imagenet64~\cite{chrabaszcz2017downsampled}.
Mixed text and image embeddings already use transformers with detection bounding boxes as input~\cite{li2020oscar}, \ie the bulk of the image processing is done in the convolutional domain.

\paragraph{The vision transformer (ViT)~\cite{dosovitskiy2020image}.} 
Interestingly, this transformer architecture is very close to the initial NLP version, devoid of explicit convolutions (just fixed-size image patch linearized into a vector), yet it competes with the state of the art for image classification. 
ViT achieves strong performance when pre-trained on a large labelled dataset such as the JFT300M   (non-public, although training on Imagenet-21k also produces competitive results).  

The need for this pre-training, in addition to strong data augmentation, can be attributed to the fact that transformers have less built-in structure than convolutions, in particular they do not have an inductive bias to focus on nearby image elements. The authors  hypothesized that a large and  varied dataset is needed to regularize the training.

In DeiT~\cite{Touvron2021DeiT}, the need for the large pre-training dataset is replaced with a student-teacher setup and stronger data augmentation and regularization, such as stochastic depth~\cite{Huang2016DeepNW} or repeated augmentation~\cite{berman2019multigrain,hoffer2020augment}. 
The teacher is a convolutional neural network that ``helps'' its student network to acquire an inductive bias for convolutions. 
The vision transformer has been thereafter successfully adopted by a wider range of computer vision tasks including object detection \cite{beal2020toward}, semantic segmentation \cite{zheng2020rethinking} and image retrieval \cite{el2021training}.

\paragraph{Positional encoding.} Transformers take a set as input, and hence are invariant to the order of the input. However, in language as well as in images, the inputs come from a structure where the order is important. The original Transformer~\cite{Vaswani2017AttentionIA} incorporates absolute non-parametric positional encoding with the input. 
Other works has replaced them with parametric encoding~\cite{ghering2017convseq2seq} or adopt Fourier-based kernelized versions~\cite{parmar2018image}. 
Absolute position encoding enforce a fixed size for the set of inputs, but some works use relative position encoding~\cite{shaw2018self} that encode the relative position between tokens. 
In our work, we replace these explicit positional encoding by positional biases that implicitly encode the spatial information.  

\paragraph{Attention along other mechanisms.}  
Several works have included attention mechanisms in neural network architectures designed for vision~\cite{wang2017residual,woo2018cbam,li2019selective,bello2019attention}.
The  mechanism is used channel-wise to capture cross-feature information that complements convolutional layers~\cite{Wang2018NonlocalNN,chen2020dynamic,zhao2020exploring}, select paths in different branch of a network~\cite{szegedy2015going}, or combine both~\cite{zhang2020resnest}. 
For instance, the squeeze-and-excite network of Hu \etal \cite{Hu2017SENet} has an attention-like module to model the channel-wise relationships between the features of a layer. 
Li \etal \cite{li2019selective} use the attention mechanism between branches of the network to adapt the receptive field of neurons.

Recently, the emergence of transformers led to hybrid architectures that benefit from other modules.
Bello~\cite{Bello2021LambdaNetworksML} proposes an approximated content attention with a positional attention component. 
Child~\etal~\cite{child2019generating} observe that many early layers in the network learn locally connected patterns, which resemble convolutions. 
This suggests that hybrid architectures inspired both by transformers and convnets are a compelling design choice. A few recent works explore this avenue for different tasks ~\cite{Srinivas2021BottleneckTF,wu2020visual}. 
In image classification, a recent work that comes out in parallel with ours is the Pyramid Vision Transformer (PVT)~\cite{wang2021pyramid}, whose design is heavily inspired by ResNet. It is principally intended to address object and instance segmentation tasks.   

Also concurrently with our work, \citet{yuan2021tokens} propose the Tokens-to-Tokens ViT (T2T-ViT) model. Similar to PVT, its design relies on re-tokenization of the output after each layer by aggregating the neighboring tokens such number of tokens are progressively reduced. 
Additionally, \citet{yuan2021tokens} investigate the integration of architecture design choices from CNNs \cite{Hu2017SENet, zagoruyko2016wide, huang2017densely} that can improve the performance and efficiency of vision transformers. 
As we will see, these recent methods are not as much focused as our work on the trade-off between accuracy and inference time. They are not competitive with respect to that compromise.

\section{Motivation}
\label{sec:motivation}

In this section we discuss the seemingly convolutional behavior of the  transformer patch projection layer. We then carry out ``grafting experiments'' of a transformer (DeiT-S) on a standard convolutional architecture (ResNet-50). The conclusions drawn by this analysis will motivate our subsequent design choices in Section~\ref{sec:model}.  

\begin{figure}[t]
\newcommand{\igconvmask}[1]{\includegraphics[width=0.45\linewidth,trim=15 15 460 0,clip]{figs/deit_conv_mask/#1.png}}
\hspace*{-1mm}
\begin{tabular}{@{}c@{\ \ \ }cc@{}}
& head 3 & head 8 \\
\raisebox{3ex}{K} & 
\igconvmask{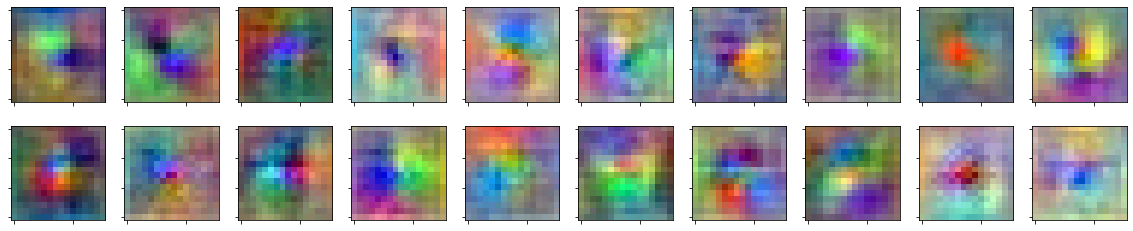} & 
\igconvmask{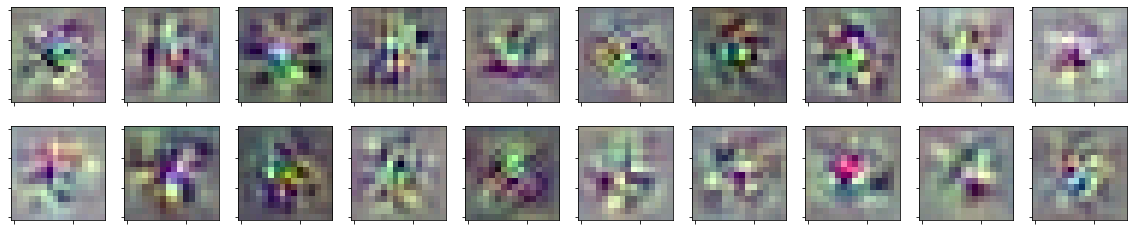} \\
\hline
\raisebox{3ex}{Q} & 
\igconvmask{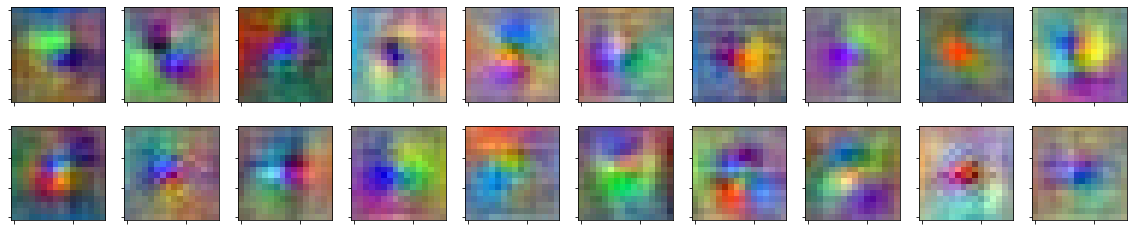} &
\igconvmask{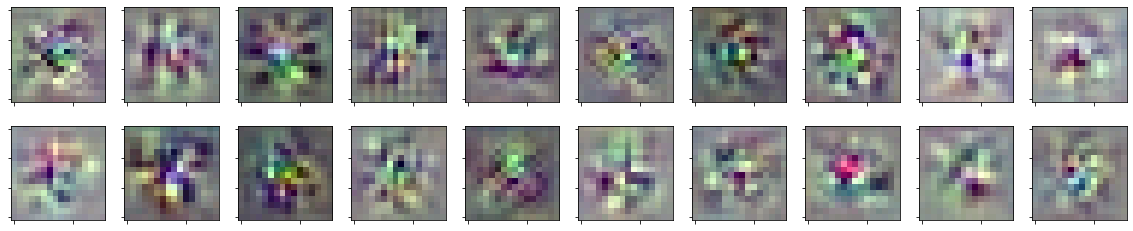} \\
\hline
\raisebox{3ex}{V} & 
\igconvmask{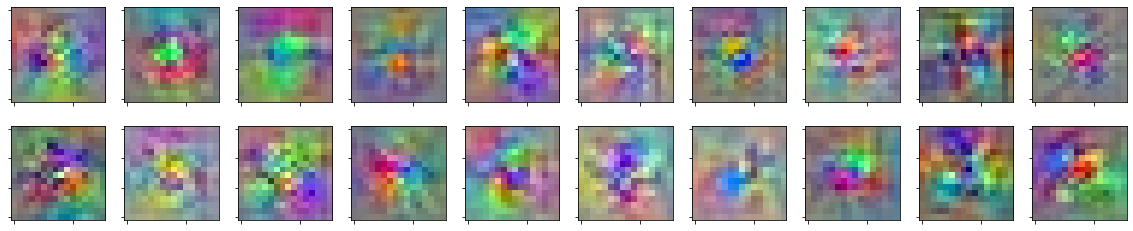} & 
\igconvmask{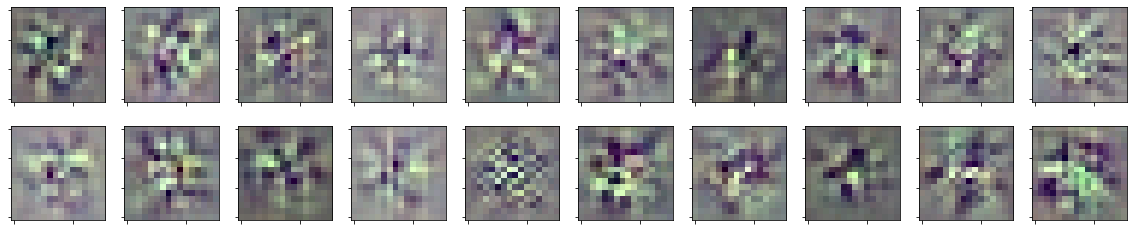} \\
\end{tabular}

\caption{\label{fig:deitfilters}
	Patch-based convolutional masks in the pre-trained DeiT-base model~\cite{Touvron2021DeiT}.
	The figure shows 12 of the 64 filters per head.
	Note that the K and Q filters are very similar, this is because the weights are entangled in the $W_\mathrm{Q} W_\mathrm{K}^\top$ multiplication. 
}
\end{figure}

\subsection{Convolutions in the ViT architecture}

ViT's patch extractor is a 16x16 convolution with stride 16. Moreover, the output of the patch extractor is multiplied by learnt weights to form the first self-attention layer's $q,k$ and $v$ embeddings, so we may consider these to also be convolutional functions of the input. This is also the case for variants like DeiT~\cite{Touvron2021DeiT} and PVT~\cite{wang2021pyramid}. In Figure~\ref{fig:deitfilters} we visualize the first layer of DeiT's attention weights, broken down by attention head.
This is a more direct representation than the principal components depicted by Dosovitskiy \etal~\cite{dosovitskiy2020image}. %
One can observe the typical patterns inherent to convolutional architectures: attention heads specialize in specific patterns (low-frequency colors / high frequency graylelvels), and the patterns are similar to Gabor filters.

In convolutions where the convolutional masks overlap significantly, the spatial smoothness of the masks comes from the overlap: nearby pixel receive approximately the same gradient.
For ViT convolutions there is no overlap. The smoothness mask is likely caused by the data augmentation: when an image is presented twice, slightly translated, the same gradient goes through each filter, so it learns this spatial smoothness.

Therefore, in spite of the absence of ``inductive bias'' in transformer architectures, the training \emph{does} produce filters that are similar to traditional convolutional layers.

\begin{figure}[t]
\includegraphics[trim=35 0 45 15,clip, width=1.04\linewidth]{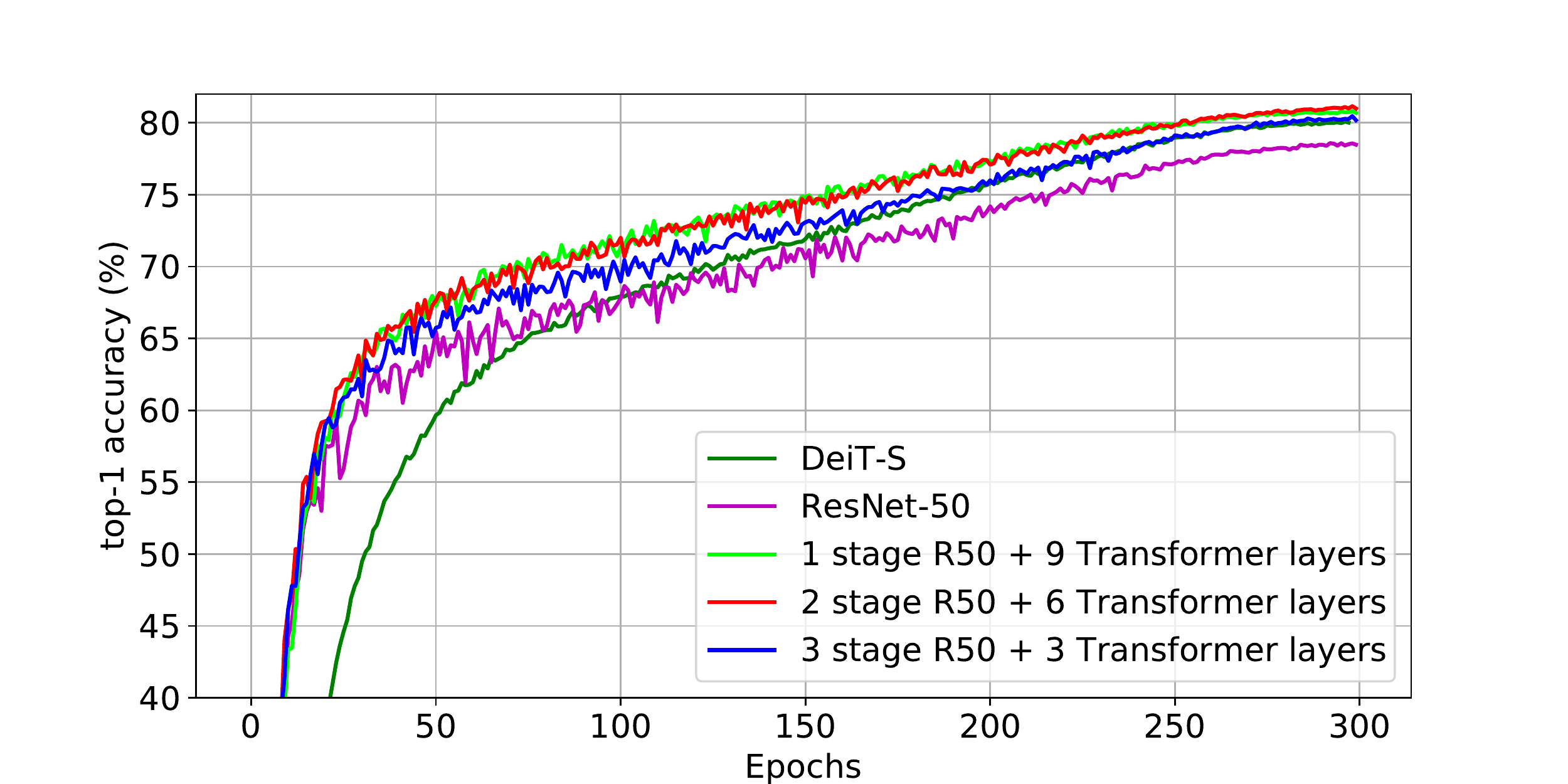}
\caption{
	\label{fig:hybrid_convergence} 
	Models with convolutional layers show a faster convergence  in the early stages
 compared to their DeiT counterpart.
}
\end{figure}

\subsection{Preliminary experiment: grafting}

The authors of the ViT image classifier~\cite{dosovitskiy2020image} experimented with stacking the transformer layers above a traditional ResNet-50. 
In that case, the ResNet acts as a feature extractor for the transformer layers and the gradients can be propagated back through the two networks.
However, in their experiments, the number of transformer layers was fixed (e.g. 12 layers for ViT-Base). 

In this subsection, we investigate the potential of mixing transformers with convolutional network \emph{under a similar computational budget}: 
We explore trade-offs obtained when varying the number of convolutional stages
and transformer layers. Our objective is to evaluate  variations of convolutional and transformer hybrids while controlling for the runtime. 

\paragraph{Grafting.}

The grafting combines a ResNet-50 and a DeiT-Small.
The two networks have similar runtimes.

We crop the upper stages of the ResNet-50 and likewise reduce the number of DeiT layers (while keeping the same number of transformer and MLP blocks). 
Since a cropped ResNet produces larger activation maps than the $14\times14$ activations consumed by DeiT, we introduce a pooling layer between them. 
In preliminary experiments we found average pooling to perform best.
The positional embedding and classification token are introduced at the interface between the convolutional and transformer layer stack. For the ResNet-50 stages, we use
ReLU activation units~\cite{nair2010rectified} and batch normalization~\cite{ioffe15batchnorm}.

\paragraph{Results.}
Table~\ref{tab:grafted} summarizes the results. 
The grafted architecture produces better results than both DeiT and ResNet-50 alone.
The smallest number of parameters and best accuracy are with two stages of ResNet-50, because this excludes the convnet's large third stage. 
Note that in this experiment, the training process is similar to DeiT: 300 epochs, we measure the top-1 validation accuracy on ImageNet, and the speed as the number of images that one GPU can process per second. 

One interesting observation that we show Figure~\ref{fig:hybrid_convergence} is that the convergence of grafted models during training seems to be similar to a convnet during the early epochs and then switch to a convergence rate similar to DeiT-S. 
A hypothesis is that the convolutional layers have the ability to learn representations of the low-level information in the earlier layers more efficiently due to their strong inductive biases, noticeably their translation invariance. 
They rely rapidly on meaningful patch embeddings, which can explain the faster convergence during the first epochs.

\paragraph{Discussion.}

It appears that in a runtime controlled regime it is beneficial to insert convolutional stages below a transformer.
Most of the processing is still done in the transformer stack for the most accurate variants of the grafted architecture. 
Thus, the priority in the next sections will be to reduce the computational cost of the transformers.
For this, instead of just grafting, the transformer architecture needs to be merged more closely with the convolutional stages.

\begin{table}[t]
\scalebox{0.8}{
\begin{tabular}{@{\ }cc|c|ccc|c@{\ }}
\toprule
\#ResNet & \#DeiT-S & nb. of  &  \multicolumn{2}{c}{FLOPs (M)} & Speed & IMNET \\
\cline{4-5} 
stages       & \ layers  & Params &  conv  & transformer  & im/s  & top-1 \\              
\midrule
0     &    12 & 22.0M & \dzo57   &  4519 & \pzo966  & 79.9 \\
1     & \pzo9 & 17.1M & \pzo820  &  3389 & \pzo995  & 80.6 \\
2     & \pzo6 & 13.1M & 1876 &  2260 & 1048     & 80.9  \\
3     & \pzo3 & 15.1M & 3385 &  1130 & 1054     & 80.1  \\
4     & \pzo0 & 25.5M & 4119 &  \tzo0  & 1254     & 78.4 \\ 
\bottomrule
\addlinespace[2mm]
\end{tabular}}
\caption{\label{tab:grafted}
	DeiT architecture grafted on top of a truncated ResNet-50 convolutional architecture.}
	\vspace{-3pt}
\end{table}

\section{Model}
\label{sec:model}

\begin{figure}
\centering
\includegraphics[width=1.0\linewidth]{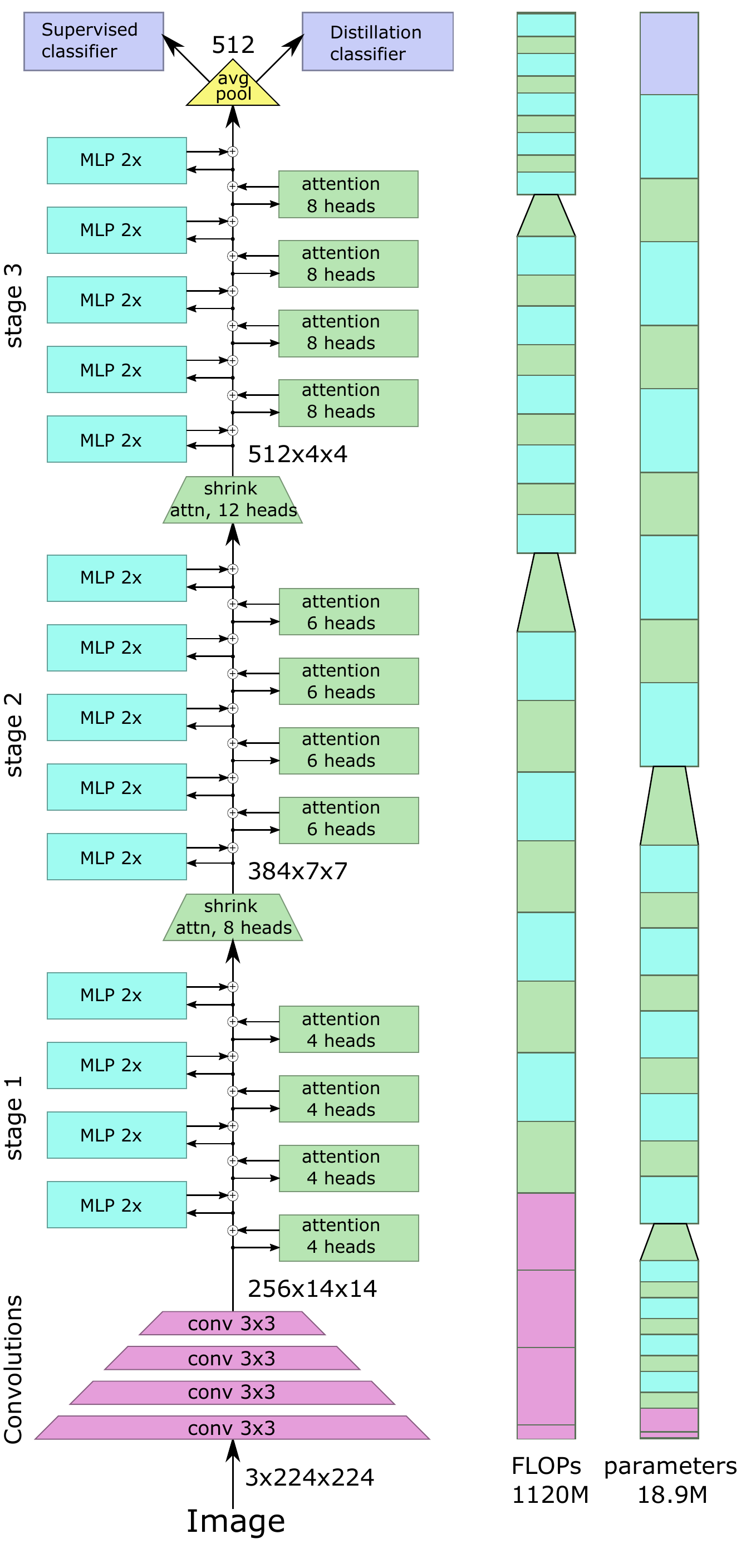}
\vspace{-15pt}
\caption{\label{fig:blockdiagram}
	Block diagram of the LeViT-256 architecture. 
	The two bars on the right indicate the relative resource consumption of each layer, measured in FLOPs, and the number of parameters.
}
\vspace{-10pt}
\end{figure}

In this section we describe the design process of the LeViT architecture and what tradeoffs were taken. 
The architecture is summarized in Figure~\ref{fig:blockdiagram}.

\subsection{Design principles of LeViT}

LeViT builds upon the ViT~\cite{dosovitskiy2020image} architecture and DeiT~\cite{Touvron2021DeiT} training method. We incorporate components that were proven useful for convolutional architectures. 
The first step is to get a compatible representation. 
Discounting the role of the classification embedding, ViT is a stack of layers that processes activation maps.
Indeed, the intermediate ``token'' embeddings can be seen as the traditional $C\times H\times W$ activation maps in FCN architectures ($BCHW$ format). 
Therefore, operations that apply to activation maps (pooling, convolutions) can be applied to the intermediate representation of DeiT. 

In this work we optimize the architecture for compute, not necessarily to minimize the number of parameters. 
One of the design decisions that makes the ResNet~\cite{He2016ResNet} family more efficient than the VGG network~\cite{Simonyan2015VGG} is to apply strong resolution reductions with a relatively small computation budget in its first two stages. 
By the time the activation map reaches the big third stage of ResNet, its resolution has already shrunk enough that the convolutions are applied to small activation maps, which reduces the computational cost. 

\subsection{LeViT components}

\paragraph{Patch embedding.}

The preliminary analysis in Section~\ref{sec:motivation} showed that the accuracy can be improved when a small convnet is applied on input to the transformer stack.
In LeViT we chose to apply 4 layers of  $3\times3$ convolutions (stride 2) to the input to perform the resolution reduction. The number of channels goes $C=3,32,64,128,256$.
This reduces the activation map input to the lower layers of the transformer 
without losing salient information.
The patch extractor for LeViT-256 transforms the image shape $(3,224,224)$ into $(256,14,14)$ with 184 MFLOPs.
For comparison, the first 10 layers of a ResNet-18 perform the same dimensionality reduction with 1042 MFLOPs.

\paragraph{No classification token.}

To use the $BCHW$ tensor format, we remove the classification token. Similar to convolutional networks, we replace it by average pooling on the last activation map, which produces an embedding used in the classifier. 
For distillation during training, we train separate heads for the classification and distillation tasks. 
At test time, we average the output from the two heads.
In practice, LeViT can be implemented using either $BNC$ or $BCHW$ tensor format, whichever is more efficient.

\paragraph{Normalization layers and activations.}

The FC layers in the ViT architecture are equivalent to $1\times1$ convolutions.
The ViT uses layer normalization before each attention and MLP unit. 
For LeViT, each convolution is followed by a batch normalization.
Following~\cite{Goyal2017AccurateLM}, each batch normalization weight parameter that joins up with a residual connection is initialized to zero.
The batch normalization can be merged with the preceding convolution for inference, which is a runtime advantage over layer normalization (for example, on EfficientNet B0, this fusion speeds up inference on GPU by a factor 2).
Whereas DeiT uses the GELU function, all of LeViT's non-linear activations are Hardswish~\cite{Howard2019SearchingFM}.

\paragraph{Multi-resolution pyramid.}

Convolutional architectures are built as pyramids, where the resolution of the activation maps decreases as their number of channels increases during processing. 
In Section~\ref{sec:motivation} we used the ResNet-50 stages to pre-process the transformer stack.

\begin{figure}
\centering
\includegraphics[width=0.43\linewidth]{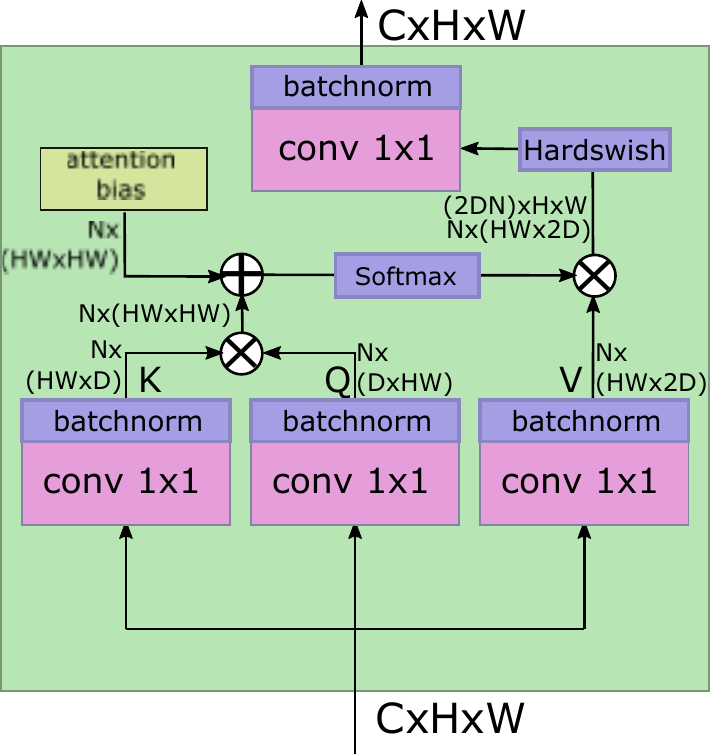} \hfill
\raisebox{0pt}{\includegraphics[width=0.54\linewidth]{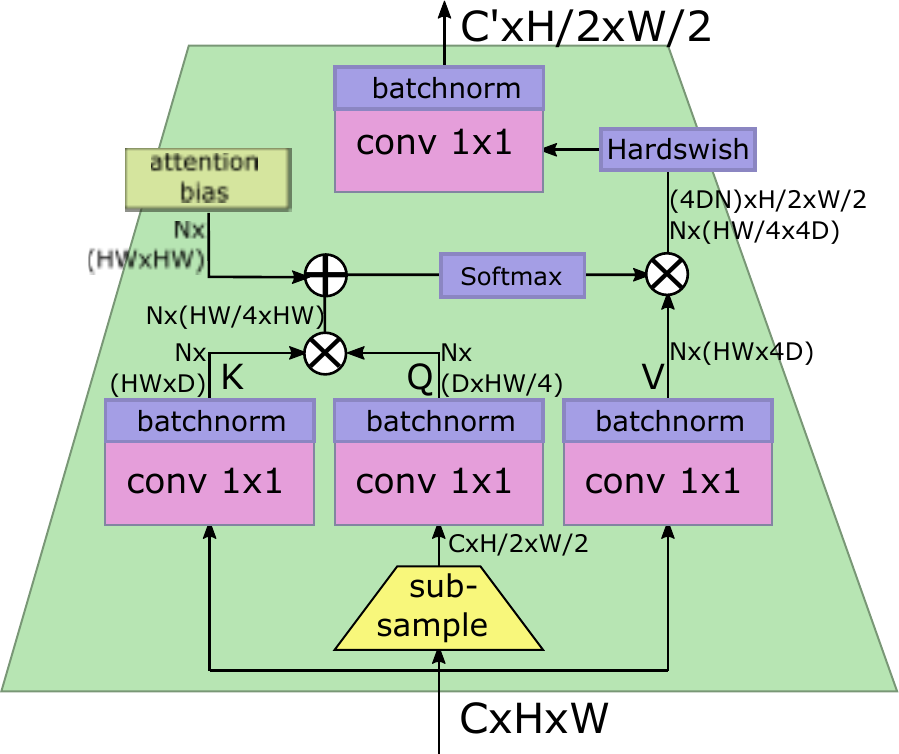}}
\caption{\label{fig:attentionblock}
	The LeViT attention blocks, using similar notations to~\cite{Wang2018NonlocalNN}.
	Left: regular version, 
	Right: with 1/2 reduction of the activation map.
	The input activation map is of size $C\times H \times W$. 
	$N$ is the number of heads, the multiplication operations are performed independently per head.
}
\end{figure}

LeViT integrates the ResNet stages within the transformer architecture. 
Inside the stages, the architecture is similar to a visual transformer: a residual structure with alternated MLP and activation blocks. 
In the following we review the modifications of the attention blocks (Figure~\ref{fig:attentionblock}) compared to the classical setup~\cite{Vaswani2017AttentionIA}. 

\paragraph{Downsampling.}

Between the LeViT stages, a \emph{shrinking attention block} reduces the size of the activation map:
a subsampling is applied before the Q transformation, which then propagates to the output of the soft activation. This maps an input tensor of size $(C,H,W)$ to an output tensor of size $(C',H/2,W/2)$ with $C'>C$. Due to the change in scale, this attention block is used without a residual connection. To prevent loss of information, we take the number of attention heads to be $C/D$.

\paragraph{Attention bias instead of a positional embedding.}

The positional embedding in transformer architectures is a location-dependent trainable parameter vector that is added to the token embeddings prior to inputting them to the transformer blocks.
If it was not there, the transformer output would be independent to permutations of the input tokens.
Ablations of the positional embedding result in a sharp drop of the classification accuracy~\cite{Chu2021PositionEncoding}. 

However positional embeddings are included only on input to the sequence of attention blocks. 
Therefore, since the positional encoding is important for higher layers as well, it is likely that it remains in the intermediate representations and needlessly uses representation capacity. 
Therefore, our goal is to provide positional information within each attention block, and to explicitly inject relative position information in the attention mechanism: 
we simply add an \emph{attention bias} to the attention maps.
The scalar attention value between two pixels $(x,y)\in [H]\times[W]$ and $(x', y')\in [H]\times[W]$ for one head $h\in[N]$ is calculated as
\begin{equation}
A^h_{(x, y), (x', y')} = Q_{(x,y),:}\bullet K_{(x',y'),:} + B^h_{|x-x'|,|y-y'|}.
\end{equation}
The first term is the classical attention. 
The second is the translation-invariant attention bias. Each head has $H\times W$ parameters corresponding to different pixel offsets. Symmetrizing the differences $x-x'$ and $y-y'$ encourages the model to train with flip invariance.

\paragraph{Smaller keys.} The bias term reduces the pressure on the keys to encode location information, so we reduce the size of the keys matrices relative to the $V$ matrix. If the keys have size $D\in\{16,32\}$, $V$ will have $2D$ channels. Restricting the size of the keys reduces the time needed to calculate the key product $QK^\top$.

For downsampling layers, where there is no residual connection, we set the dimension of $V$ to $4D$ to prevent loss of information. 

\paragraph{Attention activation.}

We apply a Hardswish activation to the product $A^h V$  before the regular linear projection is used to combine the output of the different heads. 
This is akin to a ResNet bottleneck residual block, in the sense that $V$ is the output of a  $1\times 1$ convolution, $A^h V$ corresponds to a spatial convolution, and the projection is another $1\times 1$ convolution.

\paragraph{Reducing the MLP blocks.}

The MLP residual block in ViT is a linear layer that increases the embedding dimension by a factor 4, applies a non-linearity and reduces it back with another non-linearity to the original embedding's dimension.
For vision architectures, the MLP is usually more expensive in terms of runtime and parameters than the attention block.
For LeViT, the ``MLP'' is a $1\times1$ convolution, followed by the usual batch normalization. 
To reduce the computational cost of that phase, we reduce the expansion factor of the convolution from 4 to 2.
One design objective is that attention and MLP blocks  consume approximately the same number of FLOPs.

\subsection{The LeViT family of models}

The LeViT models can spawn a range of speed-accuracy tradeoffs by varying the size of the computation stages. 
We identify them by the number of channels input to the first transformer, \eg LeViT-256 has 256 channels on input of the transformer stage. 
Table~\ref{tab:levitvariants} shows how the stages are designed for the models that we evaluate in this paper.

\def \mysp {\hspace{9pt}}
\begin{table*}
\newcommand{\ntimes}[3]{#1$\times \begin{bmatrix} C\mathrm{=}#2 \\ N\mathrm{=}#3 \end{bmatrix}$}
\newcommand{\subsamrow}[1]{\phantom{4$\times$}$\begin{bmatrix} \pzo N\mathrm{=}#1\pzo  \end{bmatrix}$}
\newcommand{\subsamrowd}[1]{\phantom{4$\times$\ }$\begin{bmatrix} \,N\mathrm{=}#1\,  \end{bmatrix}$}
\ifarxiv
\centering
\else
\scalebox{0.8}{
\fi
\begin{tabular}{|@{\mysp}l@{}|@{\mysp}c@{\mysp}|@{\mysp}c@{\mysp}|@{\mysp}c@{\mysp}|@{\mysp}c@{\mysp}|@{\mysp}c|@{\ }}
\toprule 
Model & LeViT-128S &  LeViT-128 & LeViT-192 & LeViT-256 & LeViT-384 \tabularnewline
 & ($D=16,p=0$) &  ($D=16,p=0$) & ($D=32,p=0$) & ($D=32,p=0$) & ($D=32,p=0.1$)\tabularnewline
\midrule
\begin{minipage}{1.2cm} Stage 1:\\ $14\times14$ \end{minipage} &
  \ntimes{2}{128}{4} & 
  \ntimes{4}{128}{4} & 
  \ntimes{4}{192}{3} & 
  \ntimes{4}{256}{4} & 
  \ntimes{4}{384}{6} \\[15pt]
Subsample \vspace{-6pt} &
  \subsamrow{8} &
  \subsamrow{8} & 
  \subsamrow{6} & 
  \subsamrow{8} & 
  \subsamrowd{12} \\[15pt]
\begin{minipage}{1.2cm} Stage 2: $7\times7$ \end{minipage} &
  \ntimes{3}{256}{6} & 
  \ntimes{4}{256}{8} & 
  \ntimes{4}{288}{5} & 
  \ntimes{4}{384}{6} & 
  \ntimes{4}{512}{9} \\[15pt]
Subsample \vspace{-6pt} &
  \subsamrowd{16} & 
  \subsamrowd{16} & 
  \subsamrowd{9} & 
  \subsamrowd{12} & 
  \subsamrowd{18}  \\[18pt]
\begin{minipage}{1.2cm} Stage 3: $4\times4$  \end{minipage} &
  \ntimes{4}{384}{8} & 
  \ntimes{4}{384}{12} & 
  \ntimes{4}{384}{6} & 
  \ntimes{4}{512}{8} & 
  \ntimes{4}{768}{12} \\[10pt]
\bottomrule
\addlinespace[2mm]
\end{tabular}
\ifnotarxiv
}
\hfill
\raisebox{1pt}{
\begin{minipage}{0.3\linewidth}
\fi
\caption{\label{tab:levitvariants}
	LeViT models. 
	Each stage consists of a number of pairs of Attention and MLP blocks. 
	$N$: number of heads, $C$: number of channels, $D$: output dimension of the Q and K operators.
	Separating the stages are shrinking attention blocks whose values of $C$, $C'$ are taken from the rows above and below respectively. Drop path with probability $p$ is applied to each residual connection.
	The value of $N$ in the stride-2 blocks is $C/D$ to make up for the lack of a residual connection. Each attention block is followed by an MLP with expansion factor two.
}
\ifnotarxiv
\end{minipage}}
\fi
\end{table*}

%

\section{Experiments}
\label{sec:experiments}

\subsection{Experimental context}

\paragraph{Datasets and evaluation.}

We model our experiments on the DeiT work, that is closest to our approach. It builds upon PyTorch~\cite{paszke2019pytorch} and the Timm library~\cite{pytorchmodels}. 
We train on the ImageNet-2012 dataset and evaluate on its validation set. 
We do not explore using more training data in this work.

\paragraph{Resource consumption.}

The generally accepted measure for inference speed is in units of multiply-add operations (aka FLOPs) because floating-point matrix multiplications and convolutions can be expressed as those. 

However, some operations, most notably non-linear activations, do not perform multiply-add operations. 
They are generally ignored in the FLOP counts (or counted as a single FLOP) because it is assumed that their cost is negligible w.r.t. the cost of higher-order matrix multiplications and convolutions.
However, for a small number of channels, the runtime of complicated activations like GELU is comparable to that of convolutions. 
Moreover, operations with the same number of FLOPs can be more or less efficient depending on the hardware and API used. 

Therefore, we additionally report raw timings on reference hardware, like recent papers~\cite{dosovitskiy2020image,shen2020global}. 
The efficiency of transformers relies almost exclusively on matrix multiplications with a large reduction dimension. 

\paragraph{Hardware.}

In this work, we run all experiments in PyTorch, thus we are dependent on the available optimizations in that API.
In an attempt to obtain more objective timings, we time the inference on three different hardware platforms, each corresponding to one use case: 
\begin{itemize}
\item
	One 16GB NVIDIA Volta GPU (peak performance is 12 TFLOP/s). 
	This is a typical training 
	accelerator.
\item 
	An Intel Xeon 6138 CPU at 2.0GHz. 
	This is a typical server in a datacenter, that performs feature extraction on streams of incoming images.
	PyTorch is well optimized for this configuration, using MKL and AVX2 instructions (16 vector registers of 256~bits each). 
\item 
	An ARM Graviton2 CPU (Amazon C6g instance). 
    It is a good model for the type of processors that mobile phones and other edge devices are running. 
    The Graviton2 has 32 cores supporting the NEON vector instruction set with 32 128-bit vector registers (NEON).
\end{itemize}

On the GPU we run timings on large image batches because that corresponds to typical use cases; following DeiT we use the maximum power-of-two batchsize that fits in memory.
On the CPU platforms, we measure inference time in a single thread, simulating a setting where several threads process separate streams of input images.

It is difficult to dissociate the impact of the hardware and software, so 
we experiment with several ways to optimize the network with standard PyTorch tools (the just-in-time compiler, different optimization profiles). 

\subsection{Training LeViT}

We use 32 GPUs that perform the 1000 training epochs in 3 to 5 days. 
This is more than the usual schedule for convolutional networks, but visual transformers require a long training, for example training DeiT for 1000 epochs improves by another 2 points of top-1 precision over 300 epochs. 
To regularize the training, we use distillation driven training, similar to DeiT. 
This means that LeViT is trained with two classification heads with a cross entropy loss. 
The first head receives supervision from the ground-truth classes, the second one from a RegNetY-16GF~\cite{Radosavovic2020RegNet} model trained on ImageNet. 
In fact, the LeViT training time is dominated by the teacher's inference time.

\begin{table*}
\centering
\ifnotarxiv
\scalebox{0.79}{
\fi
\begin{tabular}{l@{\mysp}r@{\mysp}r|r@{\mysp}r@{\mysp}r@{\mysp}r|r@{\mysp}r}
\toprule
              & \# params & FLOPs & \multicolumn{4}{c|}{inference speed} & \multicolumn{2}{c}{ImageNet} \\
              
                            &      &       & top-1& GPU   & Intel  & ARM  & -Real & -V2. \\
Architecture                &  (M) &   (M) & \%   & im/s  & im/s   & im/s & \%    & \%   \\
\midrule
LeViT-128S  {\bf  (ours)}   & 7.8  &  305  & 76.6 & 12880 &  131.1 & 39.1 & 83.1  & 64.3 \\
EfficientNet B0             & 5.3  &  390  & 77.1 &  4754 &   30.1 & 3.5 & 83.5  & 64.3 \\
LeViT-128 {\bf  (ours)}     & 9.2  &  406  & 78.6 &  9266 &   94.0 & 30.8 & 84.7  & 66.6 \\
\midrule
LeViT-192 {\bf (ours)}      & 10.9 &  658  & 80.0 &  8601 &   65.0 & 24.2 & 85.7  & 68.0 \\
EfficientNet B1             &  7.8 &  700  & 79.1 &  2882 &   20.0 & 2.3 & 84.9  & 66.9 \\
EfficientNet B2             &  9.2 & 1000  & 80.1 &  2149 &   13.1 & 1.3 & 85.9  & 68.8 \\
\midrule
LeViT-256 {\bf (ours)}      & 18.9 & 1120  & 81.6 &  6582 &   42.5 & 16.4 & 86.8  & 70.0 \\
DeiT-Tiny                   &  5.9 & 1220  & 76.6 &  3973 &   39.1 & 16.8 & 83.9  & 65.4 \\
EfficientNet B3             & 12   & 1800  & 81.6 &  1272 &    5.9 & 0.8 & 86.8  & 70.6 \\
\midrule
LeViT-384 {\bf (ours)}      & 39.1 & 2353  & 82.6 &  4165 &   23.1 & 9.4 & 87.6  & 71.3 \\
EfficientNet B4             & 19   & 4200  & 82.9 &   606 &    2.5 & 0.5 & 88.0  & 72.3 \\
DeiT-Small                  & 22.5 & 4522  & 82.6 &  1931 &   13.7 & 7.6 & 87.8  & 71.7 \\
\bottomrule 
\end{tabular}
\ifnotarxiv
}
\hfill
\raisebox{0pt}{
\begin{minipage}{0.25\linewidth}
\fi
\caption{\label{tab:runtime}
    Characteristics of LeViT \wrt two strong families of competitors: DeiT~\cite{Touvron2021DeiT} and EfficientNet~\cite{tan2019efficientnet}.
    The top-1 numbers are accuracies on ImageNet or ImageNet-Real and ImageNet-V2 (two last columns). 
    The others are images per second on the different platforms. 
    LeViT models optimize the trade-off between efficiency and accuracy (and not \#params). 
    The rows are sorted by FLOP counts.
}
\ifnotarxiv
\end{minipage}}
\vspace{-5pt}
\fi
\end{table*}

\def \mysp {\hspace{10pt}}

\subsection{Speed-accuracy tradeoffs}
\label{sec:speedprec}

Table~\ref{tab:runtime} shows the speed-precision tradeoffs that we obtain with LeViT, and a few salient numbers are plotted in Figure~\ref{fig:mainspeedaccuracyplot}.
We compare these with two competitive architectures from the state of the art: 
EfficientNet~\cite{tan2019efficientnet} as a strong convolutional baseline, and 
likewise DeiT~\cite{Touvron2021DeiT} a strong transformer-only architecture. 
Both baselines are trained under to maximize their accuracy.
For example, we compare with DeiT trained during 1000 epochs.

In the range of operating points we consider, the LeViT architecture largely outperforms both the transformer and convolutional variants. 
LeViT-384 is on-par with DeiT-Small in accuracy but uses half the number of FLOPs. 
The gap widens for faster operating points: LeViT-128S is on-par with DeiT-Tiny and uses 4$\times$ fewer FLOPs.

The runtime measurements follow closely these trends. 
For example LeViT-192 and LeViT-256 have about the same accuracies as EfficientNet B2 and B3 but are 5$\times$ and 7$\times$ faster on CPU, respectively. 
On the ARM platform, the float32 operations are not as well optimized compared to Intel. 
However, the speed-accuracy trade-off remains in LeViT's favor.

\begin{table}[t]
\centering
\scalebox{0.85}{
\begin{tabular}{l|cr|c}
\toprule
\quad Architecture & \#params & FLOPs & INET top-1 \\
\midrule
T2T-ViTt-14~\cite{yuan2021tokens}    
               & 21.5M  & 5200M & 80.7 \\
T2T-ViTt-19    & 39.0M  & 8400M & 81.4 \\
T2T-ViTt-24    & 64.1M  & 13200M    & 82.2 \\
BoT-S1-50~\cite{Srinivas2021BottleneckTF} 
               & 20.8M  & 4270M     & 79.1 \\
VT-R34~\cite{wu2020visual} 
               &  19.2M & 3236M & 79.9 \\
VT-R50         &  21.4M & 3412M & 80.6 \\
VT-R101        &  41.5M & 7129M & 82.3 \\
PiT-Ti~\cite{heo2021rethinking} 
              & 4.9M   & 700M & 74.6 \\
PiT-XS        & 10.6M  & 1400M & 79.1 \\
PiT-S         & 23.5M  & 2900M & 81.9 \\      
CvT-13-NAS~\cite{wu2021cvt}
              & 18M    & 4100M & 82.2 \\
\bottomrule
\end{tabular}}
\smallskip
\caption{\label{tab:sota}
    Comparison with the recent state of the art in the high-throughput regime.
	All inference are performed on images of size 224$\times$224, and training is done on ImageNet only.
}
\vspace{-10pt}
\end{table}

%
%

\subsection{Comparison with the state of the art}

Table~\ref{tab:sota} reports results with other transformer based architectures for comparison with LeViT (Table~\ref{tab:runtime}). 
Since our approach specializes in the high-throughput regime, we do not include very large and slow models~\cite{Brock2021HighPerformanceLI,tan2021efficientnetv2}. %

We compare in the FLOPs-accuracy tradeoff, since the other works are very recent and do not necessarily provide reference models on which we can time the inference. 
All Token-to-token ViT~\cite{yuan2021tokens} variants take around 5$\times$ more FLOPs than LeViT-384 and more parameters for comparable accuracies than LeViT.
Bottleneck transformers~\cite{Srinivas2021BottleneckTF} and ``Visual Transformers''~\cite{wu2020visual} (not to be confused with ViT) are both generic architectures that can also be used for detection and object segmentation. 
Both are about 5$\times$ slower than LeViT-192 at a comparable accuracy.
The same holds for the pyramid vision transformer~\cite{wang2021pyramid} (not reported in the table) but its design objectives are different.
The advantage of LeViT compared to these architectures is that it benefited from the DeiT-like distillation, which makes it much more accurate when training on ImageNet alone.
Two architecture that comes close to LeViT are the pooling-based vision transformer (PiT)~\cite{heo2021rethinking} and CvT~\cite{wu2021cvt}, ViT variants with a pyramid structure. 
PiT, the most promising one, incorporates many of the optimization ingredients for DeiT but is still 1.2$\times$ to 2.4$\times$ slower than LeViT.

\paragraph{Alternaltive evaluations.}

In Table~\ref{tab:runtime} we evaluate LeViT on alternative test sets, Imagenet Real~\cite{Beyer2020ImageNetReal} and Imagenet V2 matched frequency~\cite{Recht2019ImageNetv2}.
The two datasets use the same set of classes and training set as ImageNet. Imagenet-Real has re-assessed labels with potentially several classes per image. Imagenet-V2 (in our case match frequency) employs a different test set.
It is interesting to measure the performance on both to verify that hyper-parameters adjustments have not led to overfitting to the validation set of ImageNet. 
Thus, we measure the classification performance on the alternative test sets for models that have equivalent accuracies on ImageNet validation.
LeViT-256 and EfficientNet B3: the LeViT variant achieves the same score on -Real, but is slightly worse (-0.6) on -V2.
LeViT-384 and DeiT-Small: LeViT is slightly worse on -Real (-0.2) and -V2 (-0.4).
Although in these evaluations LeViT is relatively slightly less accurate, the speed-accuracy trade-offs still hold, compared to EfficientNet and DeiT.

\subsection{Ablations}

\begin{table}
\centering {\scalebox{0.85}
{\begin{tabular}{@{\ }l@{}l|cc|c}
\toprule 
\#id$\downarrow$ & \ \ \ Ablation of LeViT-128S     & \#params& FLOPs  & INET top-1  \tabularnewline
\midrule 
   & Base model                  & 7.4M & 305M & 71.9\\
A1 & -- without pyramid shape   & 1.2M & 308M & 56.5 \\ 
A2 & -- without PatchConv        & 7.4M & 275M & 65.3\\
A3 & -- without BatchNorm        & 7.4M & 305M & 66.6\\
A4 & -- without distillation     & 7.4M & 305M & 69.7\\
A5 & -- without attention bias   & 7.4M & 305M & 70.4\\
A6 & -- without wider blocks     & 6.2M & 312M & 70.9\\
A7 & -- without attention activ. & 7.4M & 305M & 71.1\\
\bottomrule
\end{tabular}}}
\smallskip

\caption{\label{tbl:ablation}   
    Ablation of various components \wrt the baseline LeViT-128S. 
    Each row is the baseline minus some LeViT component (1st column: experiment id).
    The training is run for 100 epochs only. 
}
\end{table}

To evaluate what contributes to the performance of LeViT, we experiment with the default setting and replace one parameter at a time.
We train the LeViT-128S model, and a number of variants, to evaluate the design changes relative to ViT/DeiT. The experiments are run with only 100 training epochs to magnify the differences and reduce training time.
The conclusions remain for larger models and longer training schedules.
We replace one component at a time, when the network needs to be reworked, we make sure the FLOP count remains roughly the same (see Appendix~\ref{sec:detailsabla} for details).  
Table~\ref{tbl:ablation} shows that all changes degrade the accuracy: 
\smallskip 

\noindent {\it A1--} The {\it without pyramid shape} ablation makes a straight stack of attention and MLPs (like DeiT). 
However, in order to keep the FLOP count similar to the baseline, the network width is reduced, resulting in a network with a small number of parameters, resulting in a very low final accuracy. 
This evidences that the reduction of the resolution in LeViT is the main tool to keep computational complexity under control.

\noindent{\it A2-- without PatchConv:} we remove the four pre-processing convolutions with a single size-16 convolution. This has little effect on the number of parameters, but the number of flops is 10\% less. The , and has a strong negative impact on the accuracy. This can be explained because in a low-capacity regime, the convolutions are an effective way to compress the $3\cdot 16^2=768$ dimensional patch input.

\noindent {\it A3--} In {\it without BatchNorm,} we replaces BatchNorm with preactivated LayerNorm, as used in the ViT/DeiT architecture. This slows down the model slightly, as batch statistics need to be calculated at test time. Removing the BatchNorm also removes the zero-initialization of the residual connections, which disrupts training.

\noindent{\it A4--} Removing the use of hard distillation from a RegNetY-16GF teacher model reduces performance, as seen with DeiT.

\noindent{\it A5--} The {\it without attention bias} ablation replaces the attention bias component with a classical positional embedding added on input to the transformer stack (like DeiT). Allowing each attention head to learn a separate bias seems to be useful.

\noindent{\it A6--} We use DeiT style blocks, i.e. $Q$,$K$ and $V$ all have dimension $D=C/N$, and the MLP blocks have expansion factor 4.

\noindent{\it A7--} LeViT has an extra Hardswish non-linearity added to the attention, in addition to the softmax non-linearity. Removing it, the {\it without attention activation} ablation degrades performance, suggesting that extra non-linearity is helpful for learning classification class boundaries.

\section{Conclusion}
\label{sec:conclusion}

This paper introduced LeViT, a transformer architecture inspired by convolutional approaches.
The accuracy of LeViT stems mainly from the training techniques in DeiT. 
Its speed comes from a series of carefully controlled design choices. 
Compared to other efficient neural nets used for feature extraction in datacenters or on mobile phones, LeViT is 1.5 to 5 times faster at comparable precision. 
Thus to the best of our knowledge, it sets a new state of the art in the trade-off between accuracy and precision in the high-speed domain.
The corresponding PyTorch code and models is available at \url{https://github.com/facebookresearch/LeViT}.

\begingroup
    \setlength{\bibsep}{5pt}
    \bibliographystyle{IEEEtranN_fullname}
    \bibliography{egbib}
\endgroup

\clearpage
\newpage

\appendix

\ifarxiv

\begin{center}
    \Huge Appendix
\end{center}

\else

\documentclass[10pt,twocolumn,letterpaper]{article}

\usepackage{iccv}
\usepackage{times}
\usepackage{epsfig}
\usepackage{graphicx}
\usepackage{amsmath}
\usepackage{amssymb}
\usepackage{booktabs}
\usepackage[numbers]{natbib}

\usepackage{balance}

\def\iccvPaperID{1103} %

\newcommand{\stdminus}[1]{\scalebox{0.65}{$\pm #1$}}
\newcommand{\rv}[1]{{\color{red}#1}}
\newcommand{\hugo}[1]{{\color{blue!20!red}[\textbf{Hugo}:#1]}}
\newcommand{\matthijs}[1]{{\color{blue}[\textbf{Matthijs}:#1]}}
\newcommand{\ben}[1]{{\color{orange}[\textbf{Benjamin}:#1]}}
\newcommand{\alaa}[1]{{\color{green!40!red}[\textbf{Alaa}:#1]}}
\newcommand{\pierre}[1]{{\color{green!20!red}[\textbf{Pierre}:#1]}}
\newcommand{\aj}[1]{{\color{green!20!red}[\textbf{Armand}:#1]}}

\def \pzo {\phantom{0}} 
\def \dzo {\phantom{00}}
\def \tzo {\phantom{000}}
\def \qzo {\phantom{0000}}

\makeatletter
\renewcommand{\paragraph}{%
  \@startsection{paragraph}{4}%
  {\z@}{0.8ex \@plus 1ex \@minus .2ex}{-1em}%
  {\normalfont\normalsize\bfseries}%
}
\makeatother

\usepackage{enumitem}
  \setlist[itemize]{%
  topsep=5pt,
  labelsep=5pt,%
  labelindent=0.4\parindent,%
  itemindent=0pt,%
  leftmargin=*,%
  itemsep=-1pt 
}

\title{    Supplementary material for \\
    LeViT: a Vision Transformer in ConvNet’s Clothing for Faster Inference
    }

\begin{document}
\maketitle
\appendix

\fi

\ifarxiv
In this appendix, we report more details and results. 
\else
In this supplementary material, we report details and results that did not fit in the main paper: 
\fi
Appendix~\ref{sec:analyses} details the timings of constituent block and provides more details about our ablation. 
We provide visualizations of the attention bias in Appendix~\ref{sec:visu}. 

\section{Detailed analysis}
\label{sec:analyses}

\subsection{Block timings}

In this section we compare the differences in design between DeiT and LeViT blocks from the perspective of a detailed runtime analysis. We measure the runtime of their constituent parts side-by-side in the supplementary Table~\ref{tab:decomptime}. For DeiT-Tiny, we replace the GELU activation with Hardswish, as otherwise it dominates the runtime.

For DeiT, we consider a block from DeiT-tiny. For LeViT, we consider a block from the first stage of LeViT-256. Both operate at resolution $14\times 14$ and have comparable run times, although LeViT is 33\% wider ($C=256$ vs $C=192$). 
Note that stage 1 is the most expensive part of  LeViT-256. In stages 2 and 3, the cost is lower due to the reduction in resolution (see Figure~4 of the main paper).

LeViT spends less time calculating the attention $QK^T$, but more time on the subsequent matrix product $AV$. Despite having the larger block width $C$, LeViT spends less time on the MLP component as the expansion factor is halved from four to two.

\begin{table}
\caption{\label{tab:decomptime}
    Timings for the components of the LeViT architecture on an Intel Xeon E5-2698 CPU core with batch size 1. 
} \smallskip

\centering \scalebox{0.85}{
\begin{tabular}{lc@{\hspace{25pt}}c}
\toprule 
Model & \multicolumn{1}{c}{DeiT-tiny} & \multicolumn{1}{c}{LeViT-256}\tabularnewline
\midrule 
Dimensions & 
$\begin{array}{c}C=192   \\ N=3\dzo \\D=64\pzo\end{array}$ 
& $\begin{array}{c}C=256 \\ N=4\dzo \\ D=32\pzo \end{array}$
\tabularnewline
\midrule 
Component & Runtime ($\mu$s) & Runtime ($\mu$s)\tabularnewline
\midrule
LayerNorm &              \dzo49  & \pzo\,n/a     \tabularnewline
Keys $Q,K$ &             \pzo299 & \pzo275 \tabularnewline
Values $V$ &             \pzo172 & \pzo275 \tabularnewline
Product $QK^{T}$ &       \pzo228 & \pzo159 \tabularnewline
Product Attention $AV$ & \pzo161 & \pzo206 \tabularnewline
Attention projection &   \pzo175 & \pzo310 \tabularnewline
MLP &                    1390    & 1140    \tabularnewline
\midrule  
Total &                  2474    & 2365    \tabularnewline
\bottomrule
\end{tabular}}
\end{table}

\subsection{More details on our ablation}
\label{sec:detailsabla}

Here we give additional details of the ablation experiments in Section 5.6 and Table~4 of the main paper. %

\paragraph{A1 -- without pyramid shape.} We test the effect of the LeViT pyramid structure, we replace the three stages with a single stage of depth 11 at resolution $14\times 14$. To preserve the FLOP count, we take $D=19$, $N=3$ and $C=2ND=114$.

\paragraph{A6 -- without wider blocks.} Compared to DeiT, LeViT blocks are relatively wide given the number of FLOPs, with smaller keys and MLP expansion factors. To test this change we modify LeViT-128S to have more traditional blocks while preserving the number of FLOPs. We therefore take $Q,K,V$ to all have dimension $D=30$, and $C=ND=120,180,240$ for the three stages. As in DeiT, the MLP expansion ratio is 4. In the subsampling layers we use $N=4C/D=16,24$, respectively.

\ifnotarxiv

\section{Alternative evaluations on Imagenet}
\label{sec:additional_results}

In Table~3 of the main paper we evaluate LeViT on alternative test sets, Imagenet Real~\cite{Beyer2020ImageNetReal} and Imagenet V2 matched frequency~\cite{Recht2019ImageNetv2}.
The two datasets use the same set of classes and training set as ImageNet. Imagenet-Real has re-assessed labels with potentially several classes per image. Imagenet-V2 (in our case match frequency) employs a different test set.
It is interesting to measure the performance on both to verify that hyper-parameters adjustments have not led to overfitting to the validation set of ImageNet. %

To evaluate this, we look at the classification performance on the alternative test sets for models that have equivalent accuracies on ImageNet validation:
\begin{itemize}
    \item LeViT-256 and EfficientNet B3: LeViT has similar accuracy on -Real, but is slightly worse (-0.6) on -V2;
    \item LeViT-384 and DeiT-Small: LeViT is slightly worse on -Real (-0.2) and -V2 (-0.4).
\end{itemize}
Although in these evaluations LeViT is relatively slightly less accurate, the speed-accuracy trade-offs still hold, compared to EfficientNet and DeiT. 

\fi

\section{Visualizations: attention bias}
\label{sec:visu}

\begin{figure*}[t]
\newcommand{\showblocks}[1]{\includegraphics[scale=0.3]{figs/attention_bias/attention_bias_#1.pdf}}
\centering \scalebox{1.2}{
{\scriptsize
    \begin{tabular}{c}
    Stage 1, attention block 1 (first one)\\[5pt]
    \showblocks{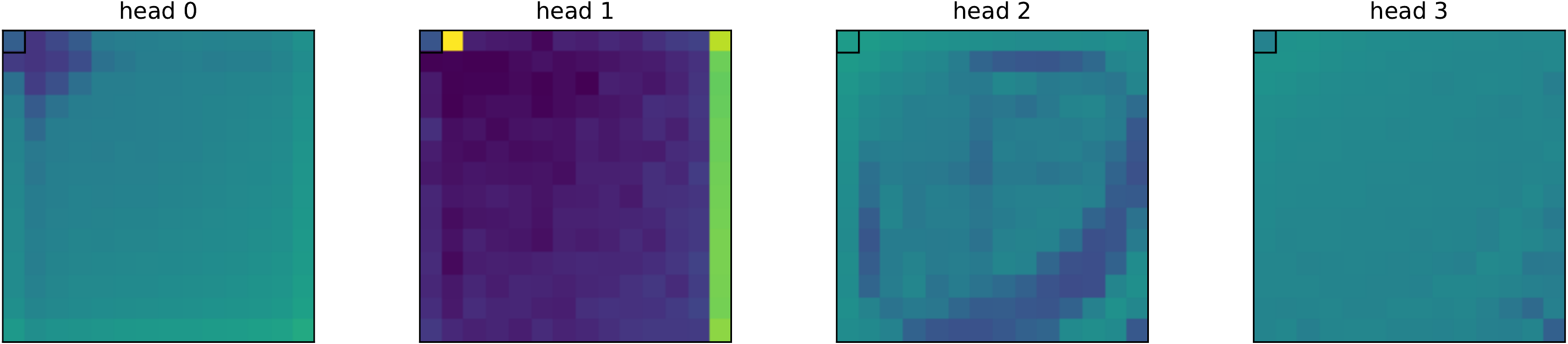} \\[5pt]
    $\vdots$ \\[5pt]
    Stage 1, attention block 4 \\[5pt]
    \showblocks{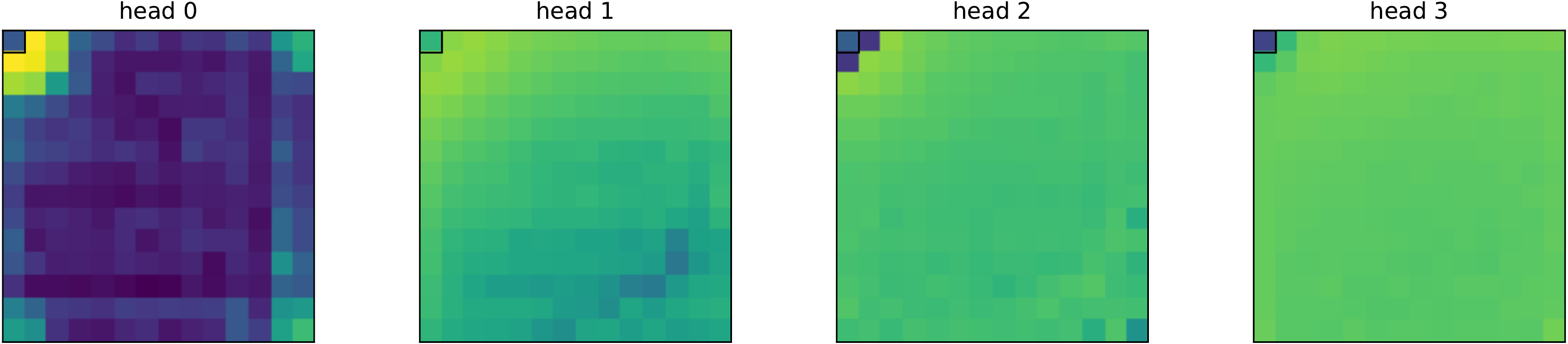} \\[8pt]
    Shrinking attention between stages 1 and 2 \\[5pt]
    \showblocks{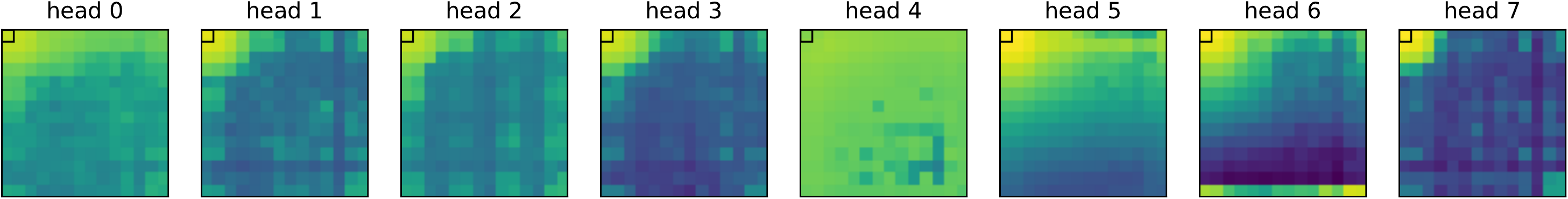} \\[8pt]
    Stage 2, attention block 1 \\[5pt]
    \showblocks{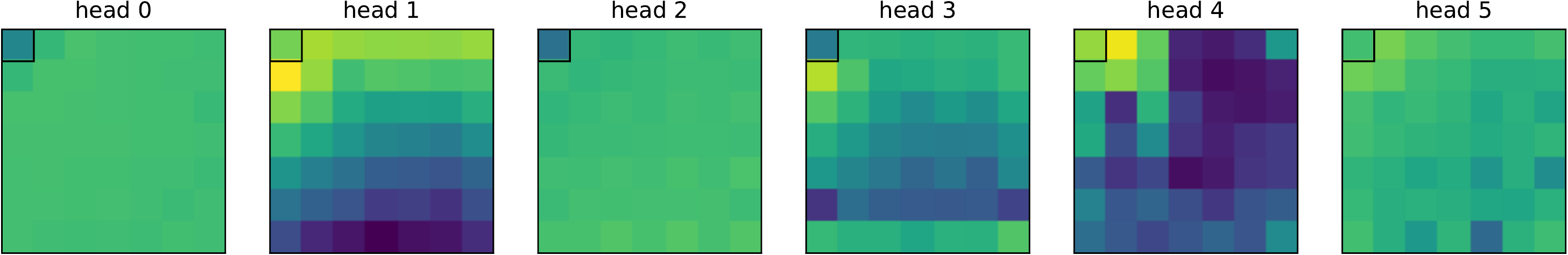} \\[5pt]
    $\vdots$ \\[5pt]
    Stage 2, attention block 4 \\[5pt]
    \showblocks{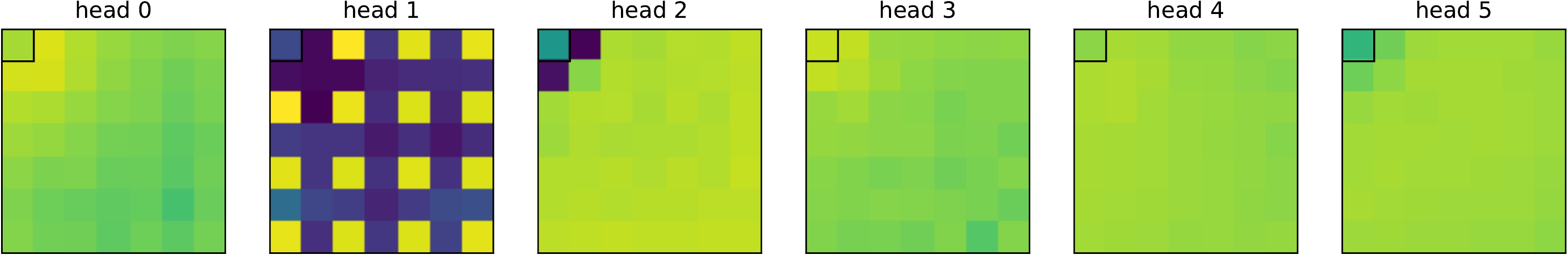} \\
    $\vdots$ \\[5pt]
    Stage 3, attention block 4 (last one) \\[5pt]
    \showblocks{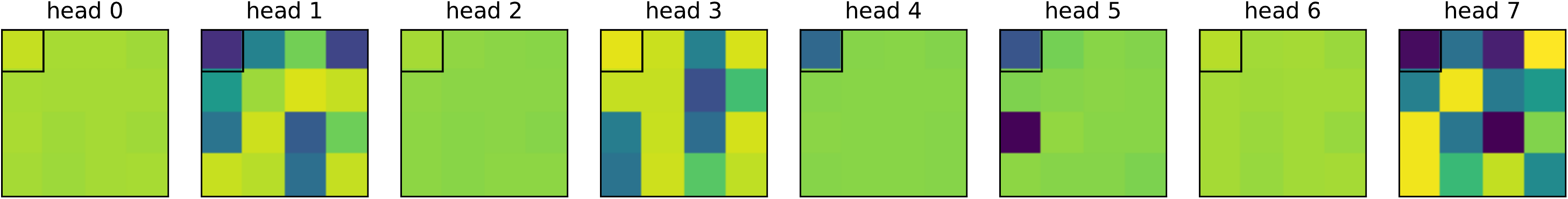} \\[5pt]
    \end{tabular}}}
    \caption{
        Visualization of the attention bias for several blocks of a trained LeViT-256 model.
        The center for which the attention is computed is the upper left pixel of the map (with a square).
        Higher bias values are in yellow, lower values in dark blue (values range from -20 to 7).
    \label{fig:attentionbias}}
\end{figure*}

The attention bias maps from Eqn.~1 in the main paper are just two-dimensional maps. Therefore we can vizualize them, see Figure~\ref{fig:attentionbias}.
They can be read as the amount of attention between two pixels that are at a certain relative position. 
The lowest values of the bias are low enough (-20) to suppress the attention between the two pixels, since they are input to a softmax.

We can observe that some heads are quite uniform, while other heads specialize in nearby pixels (\eg most heads of the shrinking attention). 
Some are clearly directional, \eg heads 1 and 4 of Stage 2/block 1 handle the pixels adjacent vertically and horizontally (respectively).
Head 1 of stage 2, block 4 has a specific period-2 pattern that may be due to the fact that its output is fed to a sub-sampling filter in the next shrinking attention block.

\ifarxiv

\else

\begingroup
  \setlength{\bibsep}{5pt}
  \bibliographystyle{IEEEtranN_fullname}
  \bibliography{egbib}
\endgroup

\end{document}

\fi

\end{document}